\documentclass[letterpaper]{article}

\usepackage{arxiv}

\usepackage[utf8]{inputenc} 
\usepackage[T1]{fontenc}    
\usepackage{hyperref}       
\usepackage{url}            
\usepackage{booktabs}       
\usepackage{amsfonts}       
\usepackage{amsmath}
\usepackage{nicefrac}       
\usepackage{microtype}      
\usepackage{lipsum}		
\usepackage{graphicx}
\usepackage{natbib}
\usepackage{doi}
\usepackage{bm}
\usepackage{float}

\title{Enhanced QKNorm normalization for neural transformers with the Lp norm}

\author{ \href{https://orcid.org/0000-0001-8231-5687}{\includegraphics[scale=0.06]{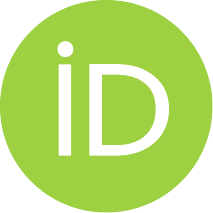}\hspace{1mm}Ezequiel L\'opez-Rubio}\thanks{Corresponding author: Ezequiel L\'opez-Rubio. Ezequiel L\'opez-Rubio and Esteban Jos\'e Palomo are also with ITIS Software. Universidad de M\'alaga. C/ Arquitecto Francisco Peñalosa 18, 29010, Málaga, Spain} \\
	Department of Computer Languages \\ and Computer Science\\
    University of M\'alaga\\
    Bulevar Louis Pasteur, 35\\
    29071 M\'alaga, Spain \\
	\texttt{ezeqlr@lcc.uma.es} \\
	\And
	\href{https://orcid.org/0000-0000-0000-0000}{\includegraphics[scale=0.06]{orcid.pdf}
    \hspace{1mm}Javier Montes-P\'erez} \\
	Department of Computer Languages \\ and Computer Science\\
    University of M\'alaga\\
    Bulevar Louis Pasteur, 35\\
    29071 M\'alaga, Spain \\
	\texttt{javimp2003uma@uma.es} \\
	\And
	\href{https://orcid.org/0000-0002-8547-9393}{\includegraphics[scale=0.06]{orcid.pdf}\hspace{1mm}Esteban Jos\'e Palomo} \\
	Department of Computer Languages \\ and Computer Science\\
    University of M\'alaga\\
    Bulevar Louis Pasteur, 35\\
    29071 M\'alaga, Spain \\
	\texttt{ejpalomo@uma.es} \\
}

\renewcommand{\shorttitle}{Enhanced QKNorm normalization for neural transformers}

\hypersetup{
pdftitle={Enhanced QKNorm normalization for neural transformers with the Lp norm},
pdfsubject={cs.AI, cs.CL},
pdfauthor={Ezequiel L\'opez-Rubio, Javier Montes-P\'erez, Esteban Jos\'e Palomo},
pdfkeywords={neural transformers, positional encoding, language models, periodic functions},
}

\begin{document}
\maketitle

\begin{abstract}
	The normalization of query and key vectors is an essential part of the Transformer architecture. It ensures that learning is stable regardless of the scale of these vectors. Some normalization approaches are available. In this preliminary work, a generalization of the QKNorm normalization scheme is proposed. The approach is based on the Lp norm, allowing non-Euclidean norms to be employed. Experimental results demonstrate the suitability of the method for a simple problem.
\end{abstract}


\section{Introduction\label{sec:Introduction}}

Transformer architectures are now the dominant models for sequence processing in natural language and other modalities, yet the underlying self-attention operation is permutation-invariant and thus has no intrinsic notion of order \citep{dufter2022position}. Beyond the question of how to encode positional information, a complementary line of work has focused on the numerical stability and optimization properties of the scaled dot-product attention mechanism \citep{vaswani2017attention}. In particular, the interaction between the scale of query and key vectors and the softmax nonlinearity has been shown to affect gradient propagation, the sharpness of attention distributions, and the robustness of training in large-scale models \citep{vaswani2017attention,qknormattention}.

A natural response to these issues is to introduce explicit normalization of the query and key representations, giving rise to Query–Key Normalization (QK-Norm) methods that normalize $Q$ and $K$ prior to computing attention scores \citep{qknormattention}. By constraining the norm of these vectors, QK-Norm decouples magnitude and direction, controls the effective dynamic range of $QK^\top$, and enables the use of more aggressive learning rates without destabilizing optimization \citep{qknormattention}. Empirical results in machine translation and language modeling indicate that such normalization can improve convergence and yield modest but consistent gains in downstream performance, all while preserving the residual structure and value pathway of standard Transformer blocks \citep{qknormattention}.

This behavior is closely related to the broader role of normalization layers in Transformer architectures. Variants such as pre-layer normalization, post-layer normalization, and more recent layouts like Peri-LN have been shown to substantially influence variance growth with depth, gradient stability, and training efficiency in billion-parameter models \citep{kim2025peri}. Peri-LN, for example, wraps each sublayer with normalization both before and after, leading to more balanced activation statistics and improved optimization compared to conventional designs \citep{kim2025peri}. At the same time, theoretical analyses suggest that Layer Normalization (LayerNorm) is not merely a stabilizing heuristic, but a key determinant of the expressive capacity of multi-head attention, as it reshapes the geometry of the function class that the model can realize \citep{brody2023expressivity}.

Taken together, these results motivate viewing attention normalization not simply as a numerical trick, but as a geometric design choice that shapes internal representations and inductive biases. Existing QK-Norm approaches, however, are almost exclusively based on Euclidean $L_2$ normalization of queries and keys \citep{qknormattention}. This implicitly fixes the underlying metric structure of the attention space and limits the degree to which one can tune the “spikiness” or entropy of attention distributions through geometric control. In this work, we therefore investigate a generalization of QK-Norm based on the $L_p$ norm, aiming to expand the design space of attention geometries while retaining the practical benefits of query–key normalization for stable and efficient Transformer training. 

The rest of the paper is organized as follows: Section \ref{sec:Methodology} describes in detail our proposal to enhance the QKNorm attention mechanism. Experiments and results are reported in Section \ref{sec:Experiments}. Section \ref{sec:Discussion} is devoted to discussing and highlighting the findings. Finally, Section \ref{sec:Conclusions} summarizes the conclusions of the paper. 

\section{Methodology\label{sec:Methodology}}

In this section, our proposal to enhance the QKNorm attention mechanism is presented. QKNorm is a modification of the scaled dot--product attention mechanism used in Transformer architectures. Introduced by \citep{qknormattention}, QKNorm stabilizes attention by explicitly normalizing query and key vectors and by replacing the fixed scaling factor with a learnable parameter. First, the standard Transformer attention is reviewed along with the QKNorm formulation (Subsection \ref{subsec:standard-qknorm}). Then, our approach is detailed (Subsection \ref{subsec:lpnorm-attention}).

\subsection{\label{subsec:standard-qknorm}Standard and QKNorm attention}

Let $\mathbf{X} \in \mathbb{R}^{n \times d}$ denote a sequence of $n$ input embeddings of dimension $d$. For a single attention head, the Transformer computes linear projections
\begin{align}
    \mathbf{Q} &= \mathbf{X}\mathbf{W}^Q, \\
    \mathbf{K} &= \mathbf{X}\mathbf{W}^K, \\
    \mathbf{V} &= \mathbf{X}\mathbf{W}^V,
\end{align}
where $\mathbf{W}^Q, \mathbf{W}^K, \mathbf{W}^V \in \mathbb{R}^{d \times d_k}$ are learned matrices and $d_k$ is the dimensionality per head.

The attention output is then computed as
\begin{equation}
    \mathrm{Attention}(\mathbf{Q},\mathbf{K},\mathbf{V})
    =
    \mathrm{softmax}\!\left(
    \frac{\mathbf{Q}\mathbf{K}^\top}{\sqrt{d_k}}
    \right)\mathbf{V}.
\end{equation}

The factor $\sqrt{d_k}$ is introduced to control the variance of the dot products and prevent saturation of the softmax function.

Despite scaling, dot-product attention can still produce large logits when query and key vectors have large norms. This can lead to unstable gradients and degraded performance, particularly in low-resource or noisy training regimes. QKNorm addresses this issue by explicitly normalizing query and key vectors to unit length, and replacing the fixed scaling factor with a learnable parameter.

Let $\mathbf{q}_i$ and $\mathbf{k}_j$ denote individual query and key vectors. QKNorm applies row--wise $\ell_2$ normalization:
\begin{align}
    \hat{\mathbf{q}}^{(2)}_{i} &= \frac{\mathbf{q}_i}{\lVert \mathbf{q}_i \rVert_2}, \\
    \hat{\mathbf{k}}^{(2)}_{j} &= \frac{\mathbf{k}_j}{\lVert \mathbf{k}_j \rVert_2}.
\end{align}

Equivalently, in matrix form:
\begin{align}
    \hat{\mathbf{Q}}^{(2)} &= \mathrm{norm}_{\ell_2}(\mathbf{Q}), \\
    \hat{\mathbf{K}}^{(2)} &= \mathrm{norm}_{\ell_2}(\mathbf{K}),
\end{align}
where normalization is applied independently to each row. This ensures that all query and key vectors lie on the unit hypersphere.

Instead of scaling by $\sqrt{d_k}$, QKNorm introduces a learnable, positive scalar parameter, $\alpha$. The attention logits are computed as
\begin{equation}
    s^{(2)}_{ij} = \alpha \, \left( {\hat{\mathbf{q}}^{(2)}_{i}} \right) ^\top \hat{\mathbf{k}}^{(2)}_{j}.
\end{equation}

Since $\left( {\hat{\mathbf{q}}^{(2)}_{i}} \right) ^\top \hat{\mathbf{k}}^{(2)}_{j}$ corresponds to cosine similarity, the logits are naturally bounded, improving numerical stability. The final attention operation is given by
\begin{equation}
    \mathrm{QKNormAttn}(\mathbf{Q},\mathbf{K},\mathbf{V})
    =
    \mathrm{softmax}(\mathbf{S}^{(2)})\,\mathbf{V},
\end{equation}
where
\begin{equation}
    \mathbf{S}^{(2)} = \alpha \, \hat{\mathbf{Q}}^{(2)} \left( \hat{\mathbf{K}}^{(2)} \right) ^\top.
\end{equation}

\subsection{\label{subsec:lpnorm-attention}Lp norm attention}

We propose to generalize the $\ell_2$ normalization of the original QKNorm approach to an $\ell_p$ normalization, where $p$ is a hyperparameter, $p\in\mathbb{R}$, with the constraint $p \ge 1$. This constraint ensures that a proper vector norm is used. Therefore, we have:

\begin{align}
    \hat{\mathbf{q}}^{(p)}_{i} &= \frac{\mathbf{q}_i}{\lVert \mathbf{q}_i \rVert_p}, \\
    \hat{\mathbf{k}}^{(p)}_{j} &= \frac{\mathbf{k}_j}{\lVert \mathbf{k}_j \rVert_p},
\end{align}

where the $\ell_p$ norm of a vector $\mathbf{v}\in\mathbb{R}^{d_k}$ is given by:

\begin{equation}
\lVert\mathbf{v}\rVert_{p}=\left(\sum_{h=1}^{d_k}\left|v_{h}\right|^{p}\right)^{\frac{1}{p}}.
\end{equation}

Next, the attention logits are obtained as
\begin{equation}
    s^{(p)}_{ij} = \alpha \, \left( {\hat{\mathbf{q}}^{(p)}_{i}} \right) ^\top \hat{\mathbf{k}}_{j}^{(p)}.\label{eq:Lp-logits}
\end{equation}

The above logits (\ref{eq:Lp-logits}) are bounded because they are obtained as the dot product of two $\ell_p$ normalized vectors, $\left( {\hat{\mathbf{q}}^{(p)}_{i}} \right) ^\top \hat{\mathbf{k}}_{j}^{(p)}$. The final attention operation is computed as follows:
\begin{equation}
    \mathrm{QKLpNormAttn}(\mathbf{Q},\mathbf{K},\mathbf{V}, p)
    =
    \mathrm{softmax}(\mathbf{S}^{(p)})\,\mathbf{V},
\end{equation}
where
\begin{equation}
    \mathbf{S}^{(p)} = \alpha \, \hat{\mathbf{Q}}^{(p)} \left( \hat{\mathbf{K}}^{(p)} \right) ^\top.
\end{equation}

\section{Experiments\label{sec:Experiments}}

We conducted our experiments using the \texttt{nanoGPT} \citep{nanogptkaparthy} decoder-only model,
which implements a simplified GPT-style Transformer \citep{vaswani2017attention}.
The model employs pre-norm residual blocks with causal self-attention and
a two-layer multilayer perceptron (MLP) in each block.
The configuration used in our experiments consists of 6 layers, 6 attention heads, an embedding dimension of 384, a dropout rate of 0.2, and a context length of 256. The vocabulary comprises 65 characters.

We evaluate the generalized QKNorm attention \citep{qknormattention} formulation in which each head's query
and key vectors are $\ell_p$-normalized prior to the dot product, as proposed in Subsection \ref{subsec:lpnorm-attention}.
The hyperparameter $p$ is varied according to the following sweep:
\begin{equation}
p \in \{1.0, 1.5, 2.0, 2.5, 3.0, 3.5, 4.0\}.
\end{equation}

The dataset used is called \texttt{Tiny Shakespeare} \citep{shakespearedataset}, a small text corpus from Hugging Face consisting of 40000 lines written by the English poet. Furthermore, we use it at the character level, meaning that the Transformer processes
individual characters as tokens rather than subword units. Raw text is encoded into integer IDs and stored for its use during training.

Specifically, the architecture is trained using $K$-fold cross-validation with $K=10$. Therefore, taking into account the seven possible values of the normalisation hyperparameter and the ten splits of the dataset, this results in a total of 70 training runs. All of them were carried out on a single NVIDIA DGX A100 GPU (40 GB VRAM) running Python 3.12, and PyTorch 2.10.0.

\begin{figure}[H]
   \centering
   \includegraphics[width=0.7\linewidth]{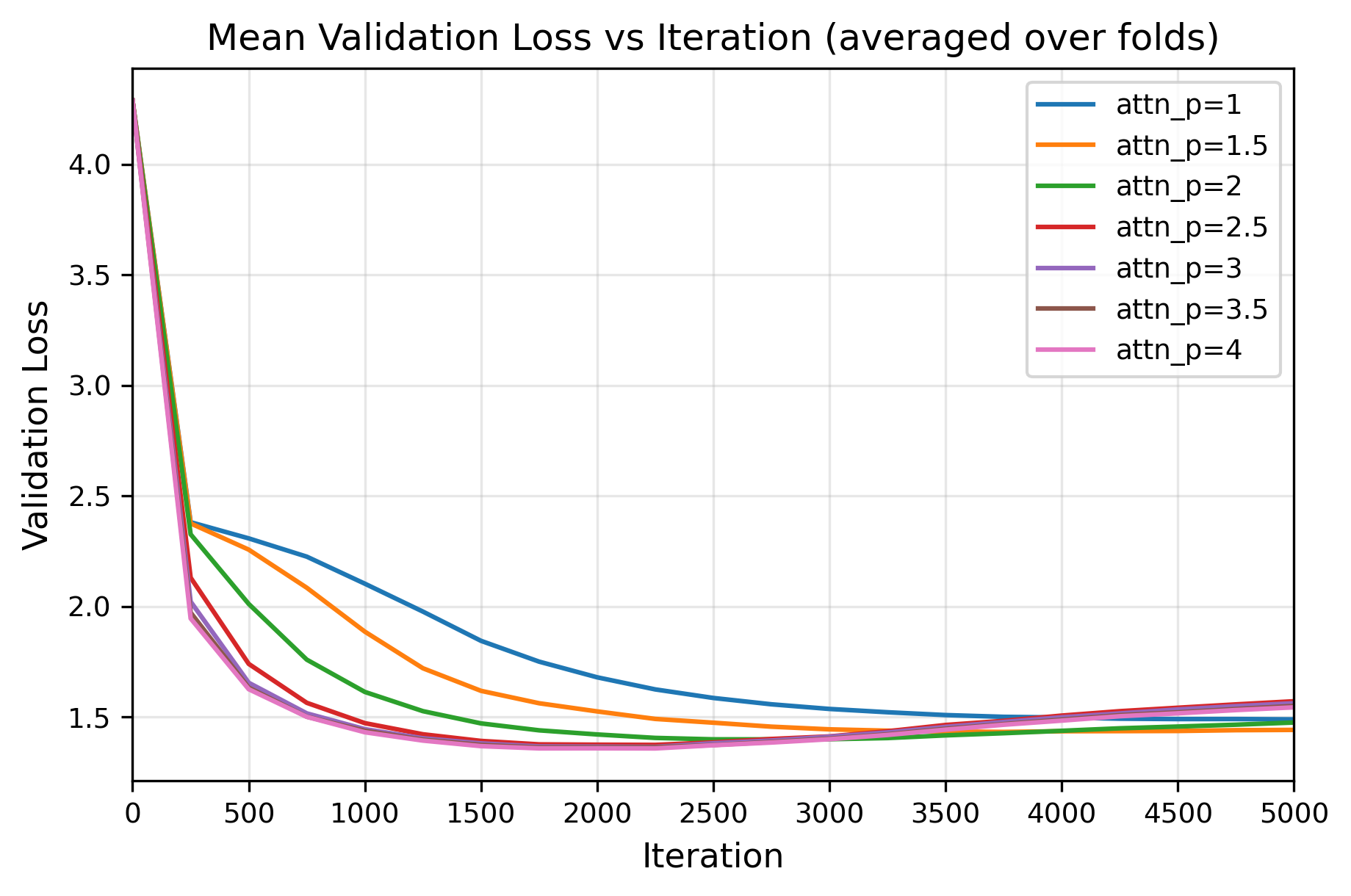}
   \caption{Per-$p$ validation loss curves, averaged by fold.}
   \label{fig:perfold_by_p}
\end{figure}

As shown in Figure~\ref{fig:perfold_by_p}, the validation cross-entropy loss is plotted as a function of the training iterations, averaged over the 10 folds. The results are highly positive, as the minimum loss is reached earlier for $p=2.5$, $p=3$, $p=3.5$, and $p=4$ compared to $p=2$, which corresponds to the default value used in standard \texttt{QKNorm}. Moreover, the attained minimum loss is lower, further supporting the proposed conclusion. While the standard approach $p=2$ gets its minimum at 1.40506, our proposal outperforms it for all tried values with $p>2$:

\begin{itemize}
  \item $p=2.5$ achieves an averaged $min$ value of 1.373479
  \item $p=3$ achieves an averaged $min$ value of 1.364543
  \item $p=3.5$ achieves an averaged $min$ value of 1.361630
  \item $p=4$ achieves an averaged $min$ value of 1.357461
\end{itemize}

These average results are further confirmed in Appendix \ref{sec:appendix-perfold}, where the validation cross-entropy loss curves are shown individually for each fold. It can be seen that the behavior is the same for all folds.

\begin{figure}[H]
   \centering
   \includegraphics[width=0.7\linewidth]{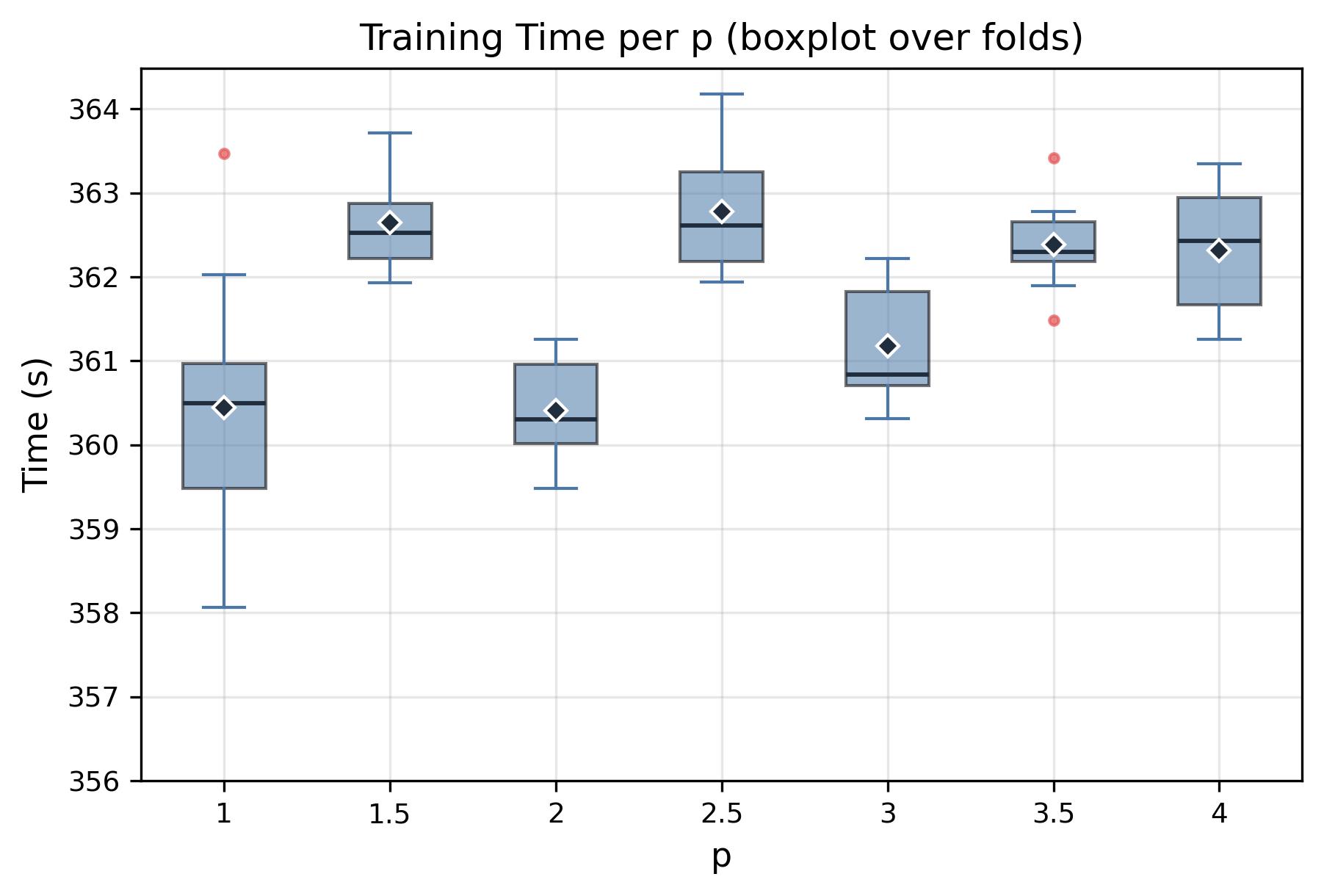}
   \caption{Per-$p$ training time distribution across folds: the central line denotes the median, the box spans the interquartile range (IQR), whiskers extend to 1.5×IQR, and outliers are shown as individual points; the diamond marks the mean.}
   \label{fig:trainingtimes}
\end{figure}

With respect to the training run durations, Figure~\ref{fig:trainingtimes}
presents a bar chart showing that the mean training time remains essentially
invariant across the tested values of $p$ (approximately 360--363\,s).
The observed variation is limited to a few seconds (less than 1\%),
which falls within the expected noise due to data ordering effects,
I/O jitter and stochastic variability across folds.
Consequently, the choice of $p$ does not introduce any meaningful
computational overhead in this setting and should therefore be guided
by validation performance rather than runtime considerations.


\section{Discussion\label{sec:Discussion}}

Next, we discuss the key features of our proposal and the most relevant aspects of the experimental results. The $\ell_p$ norm varies continuously according to the value of $p$. The higher the value of $p$, the more importance is given to the vector components with a larger absolute value. In the limit $p \to \inf$, also called the maximum metric, only the vector component with the largest absolute value is considered. In the context of the attention mechanism, this means that larger values of $p$ correspond to greater attention to vector components with the highest absolute values. So the Transformer architecture would focus on fewer features as $p$ increases. Therefore, the proposed $p$ hyperparameter controls the span of features (vector components) deemed relevant.

As seen in the experiments, the Transformer model's behavior varies continuously as $p$ is changed. This is consistent with the continuous variation of the $\ell_p$ norm and demonstrates a stable change in the model's behavior as $p$ varies. Moreover, Figure~\ref{fig:perfold_by_p} shows that the learning process is also stable over time for all values tested of $p$, with a smooth progress of the validation loss. The performance curves indicate that values of $p$ above the standard $p=2$ yield the best validation loss. Regarding computational load (Figure~\ref {fig:trainingtimes}), our approach incurs no overhead compared to the original QKNorm. Furthermore, for $p > 2$, the learning process attains its minimum (validation loss) earlier, i.e., these values of $p$ lead to faster validation loss minimization. It is also observed that the results converge as $p$ grows. 

The reported results of the computational experiments suggest that Euclidean normalization ($p=2$) may not be necessary for the attention mechanism to operate successfully. Maybe reducing the importance given to vector components with smaller absolute values is the driver for better performance, as shown for $p > 2$.

\section{Conclusions\label{sec:Conclusions}}

A new methodology for the attention mechanism of Transformer neural architectures has been proposed. The methodology is a generalization of the QKNorm normalization scheme. The proposal is based on the $\ell_p$ norm. The newly introduced hyperparameter $p$ controls the span of features considered by the attention mechanism. Experimental results demonstrate that our approach outperforms the original QKNorm in both validation loss and convergence speed.

\clearpage
\appendix
\section{Per-fold validation loss curves\label{sec:appendix-perfold}}

This appendix includes the validation loss curves for each of the $K=10$ folds used in the cross-validation experiments, to facilitate inspection of fold-to-fold consistency.

\begin{figure}[H]
  \centering

  \begin{minipage}[t]{0.24\linewidth}
    \centering
    \IfFileExists{images/fold_0_attn_p_comparison.png}{\includegraphics[width=\linewidth]{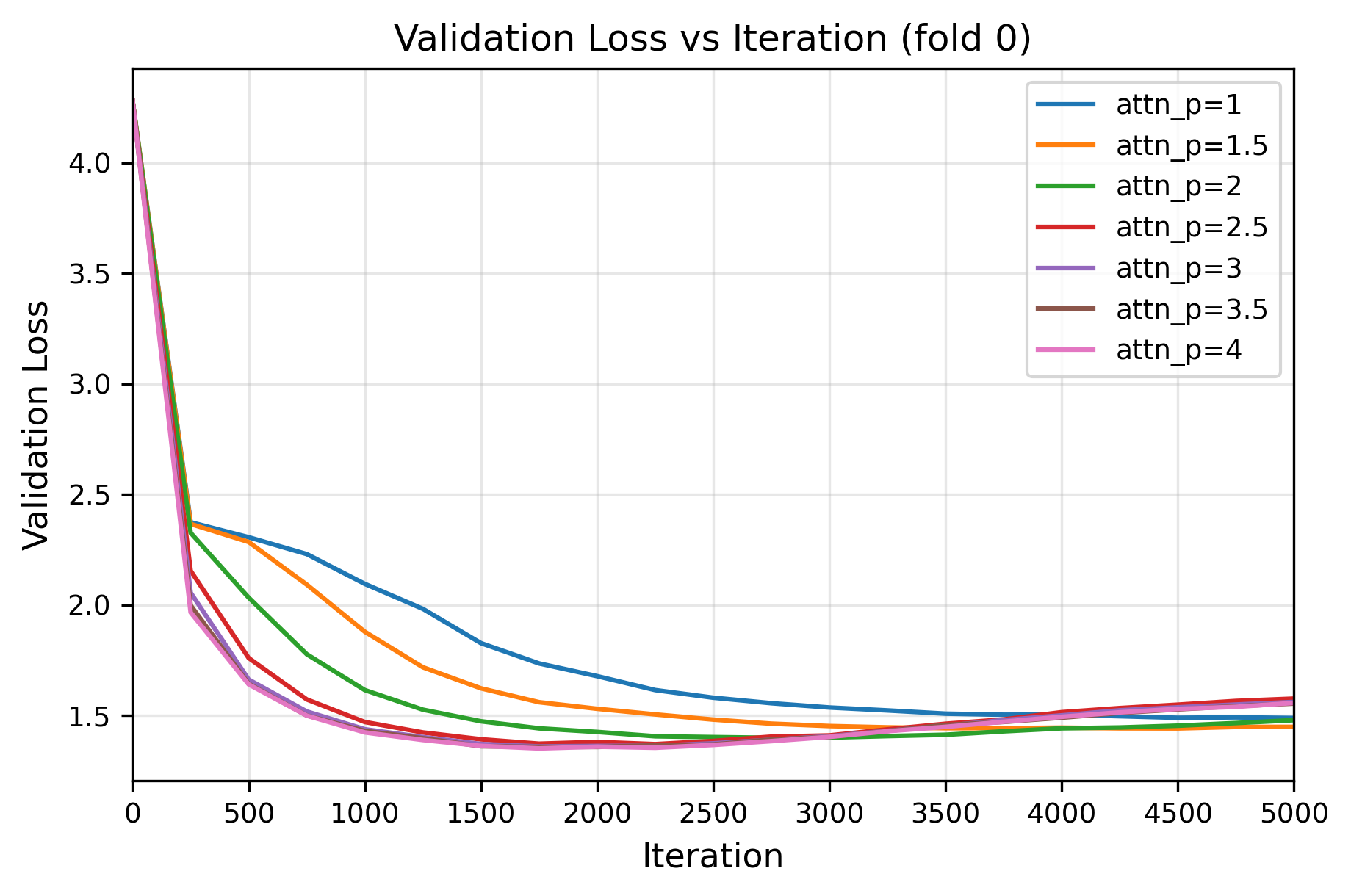}}{\fbox{\scriptsize Missing: fold\_0}}
    {\scriptsize Fold 0}
  \end{minipage}\hfill
  \begin{minipage}[t]{0.24\linewidth}
    \centering
    \IfFileExists{images/fold_1_attn_p_comparison.png}{\includegraphics[width=\linewidth]{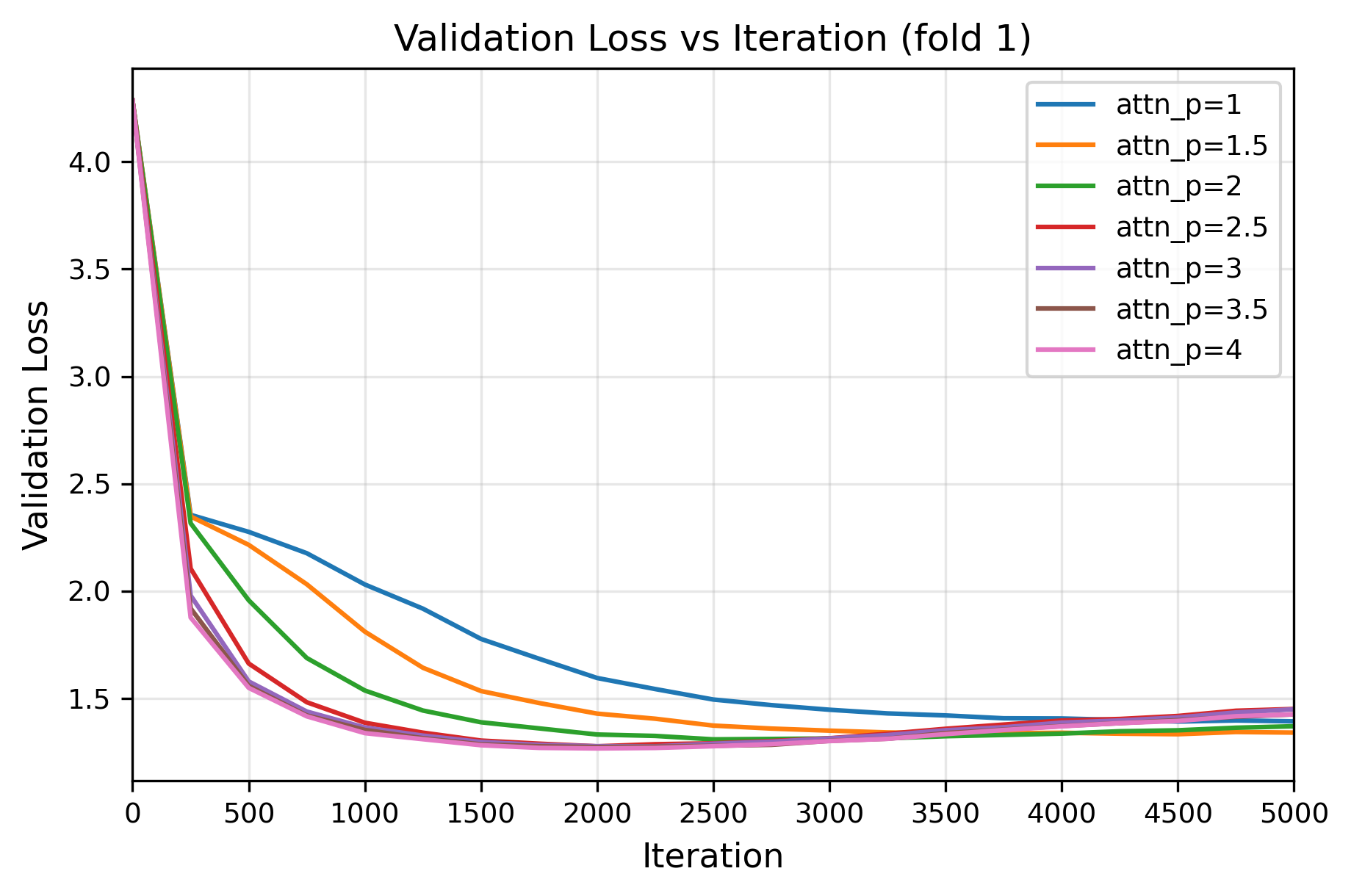}}{\fbox{\scriptsize Missing: fold\_1}}
    {\scriptsize Fold 1}
  \end{minipage}\hfill
  \begin{minipage}[t]{0.24\linewidth}
    \centering
    \IfFileExists{images/fold_2_attn_p_comparison.png}{\includegraphics[width=\linewidth]{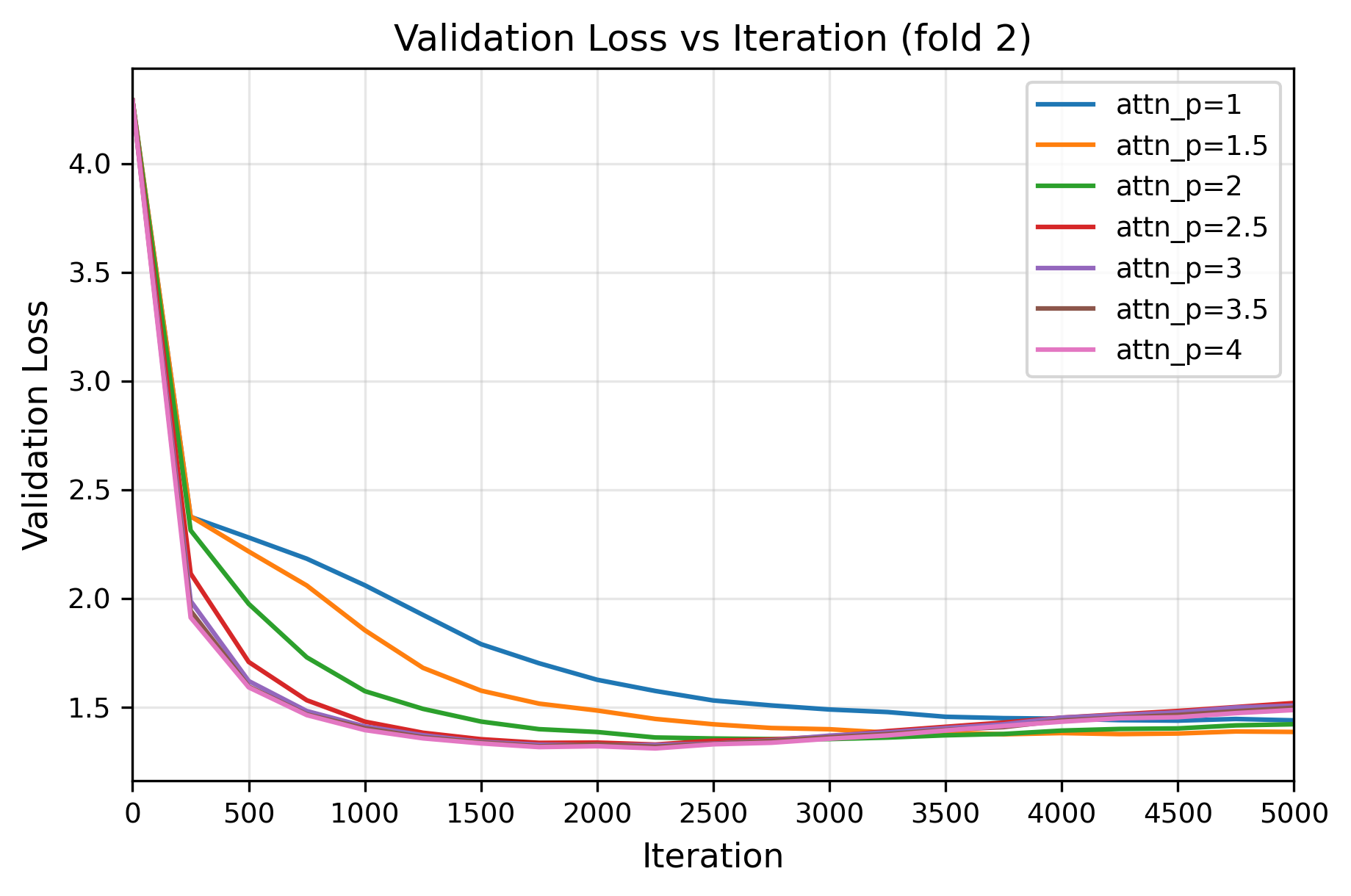}}{\fbox{\scriptsize Missing: fold\_2}}
    {\scriptsize Fold 2}
  \end{minipage}\hfill
  \begin{minipage}[t]{0.24\linewidth}
    \centering
    \IfFileExists{images/fold_3_attn_p_comparison.png}{\includegraphics[width=\linewidth]{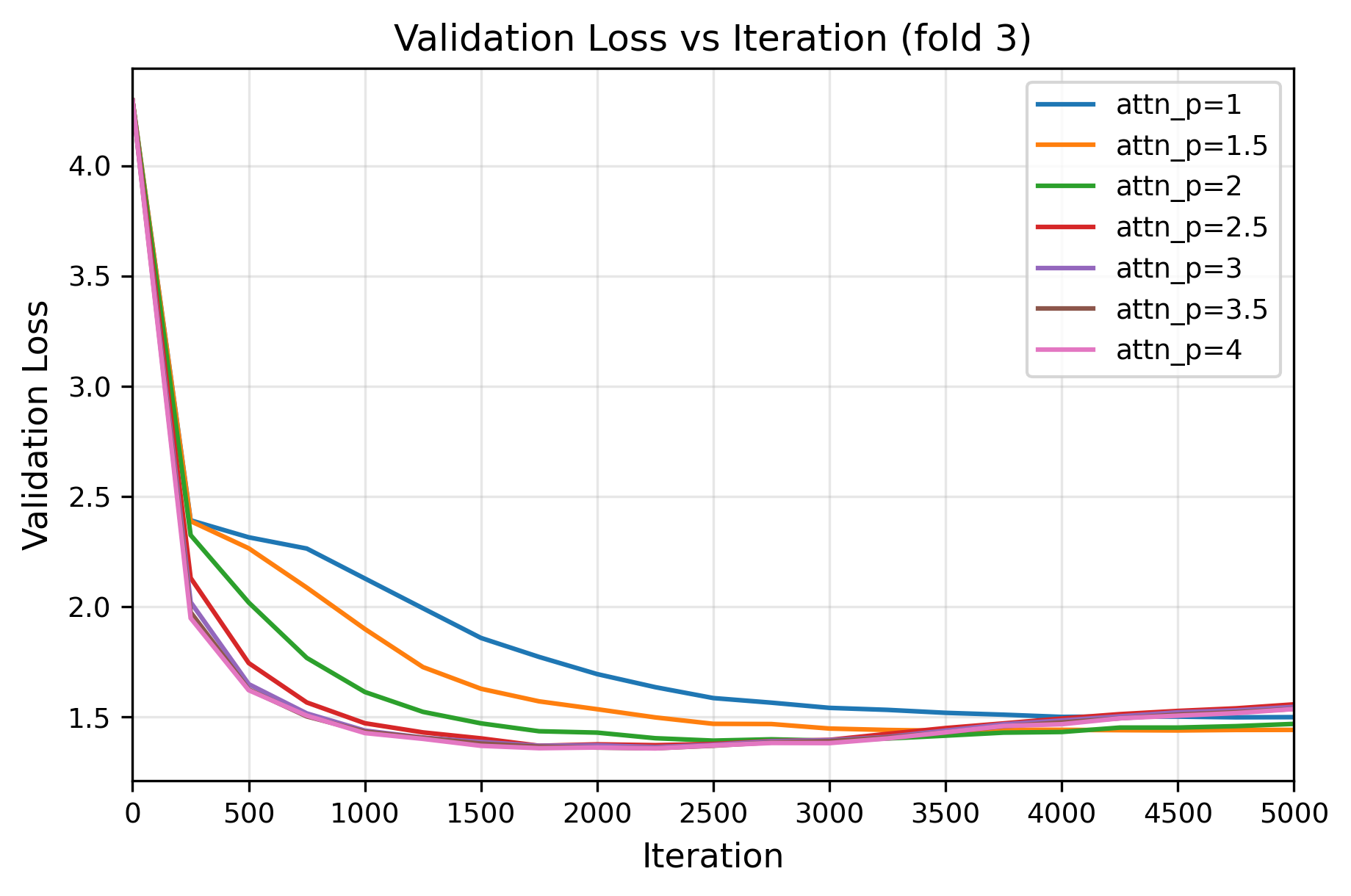}}{\fbox{\scriptsize Missing: fold\_3}}
    {\scriptsize Fold 3}
  \end{minipage}

  \vspace{0.8em}

  \begin{minipage}[t]{0.24\linewidth}
    \centering
    \IfFileExists{images/fold_4_attn_p_comparison.png}{\includegraphics[width=\linewidth]{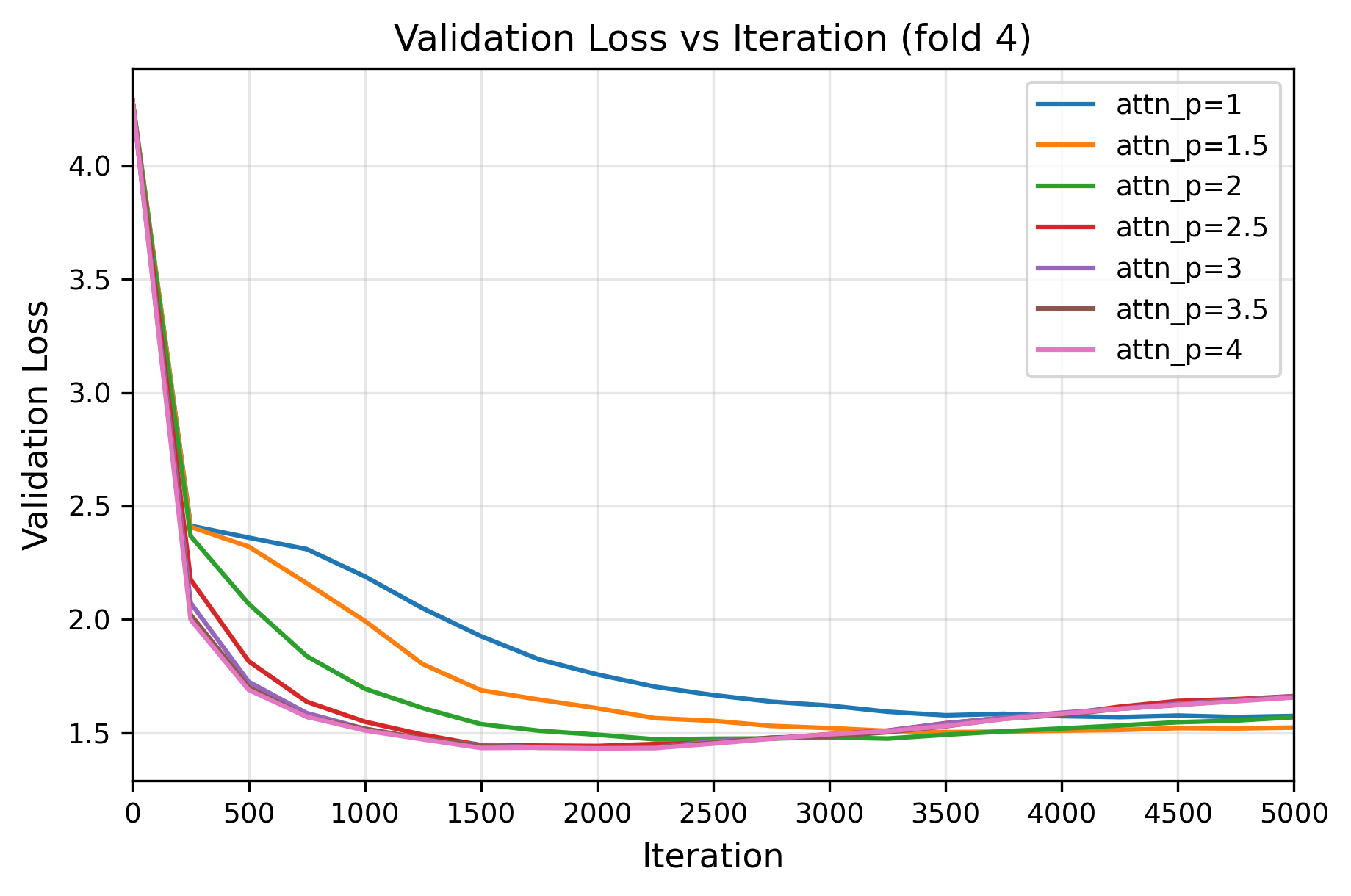}}{\fbox{\scriptsize Missing: fold\_4}}
    {\scriptsize Fold 4}
  \end{minipage}\hfill
  \begin{minipage}[t]{0.24\linewidth}
    \centering
    \IfFileExists{images/fold_5_attn_p_comparison.png}{\includegraphics[width=\linewidth]{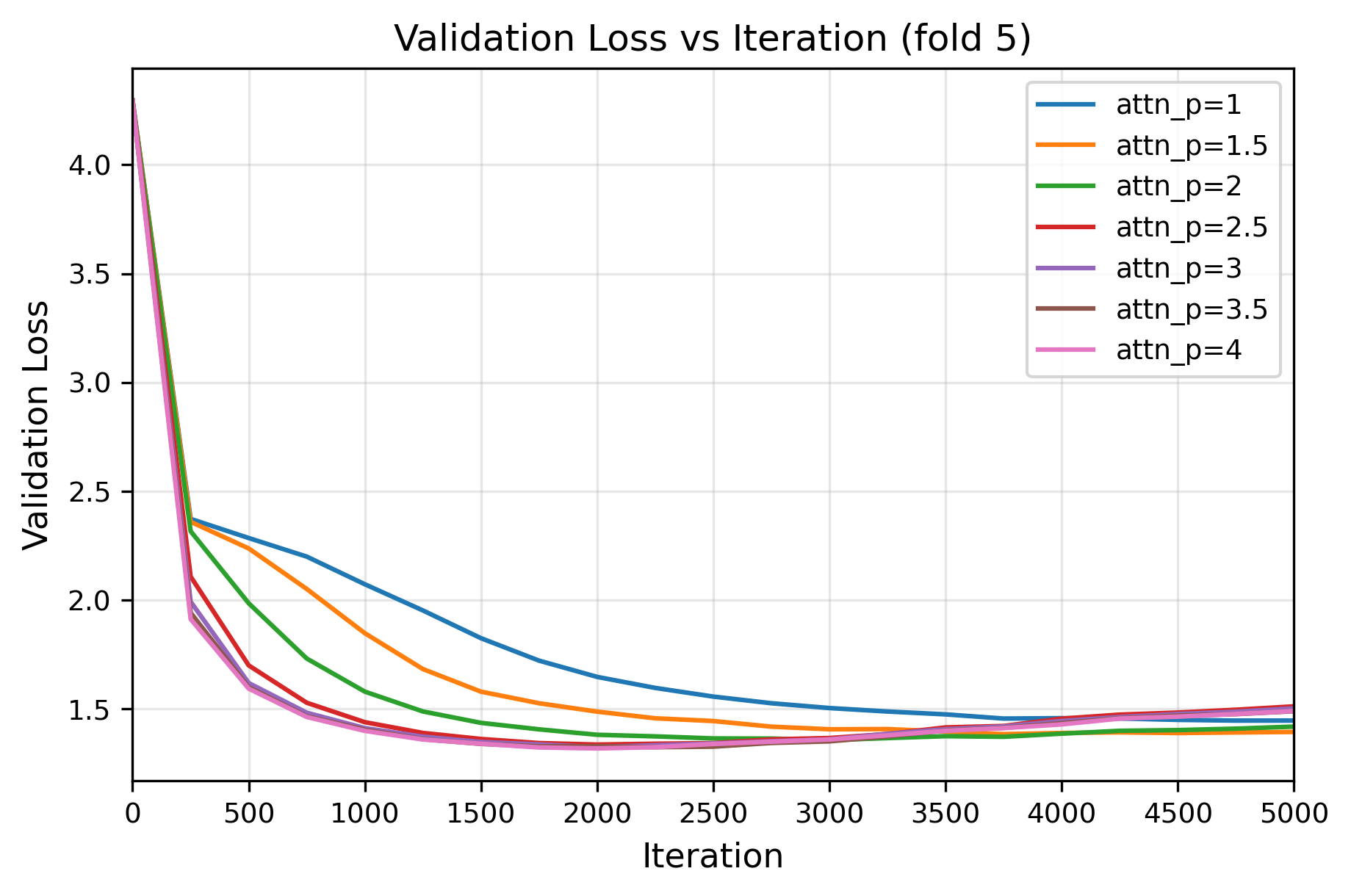}}{\fbox{\scriptsize Missing: fold\_5}}
    {\scriptsize Fold 5}
  \end{minipage}\hfill
  \begin{minipage}[t]{0.24\linewidth}
    \centering
    \IfFileExists{images/fold_6_attn_p_comparison.png}{\includegraphics[width=\linewidth]{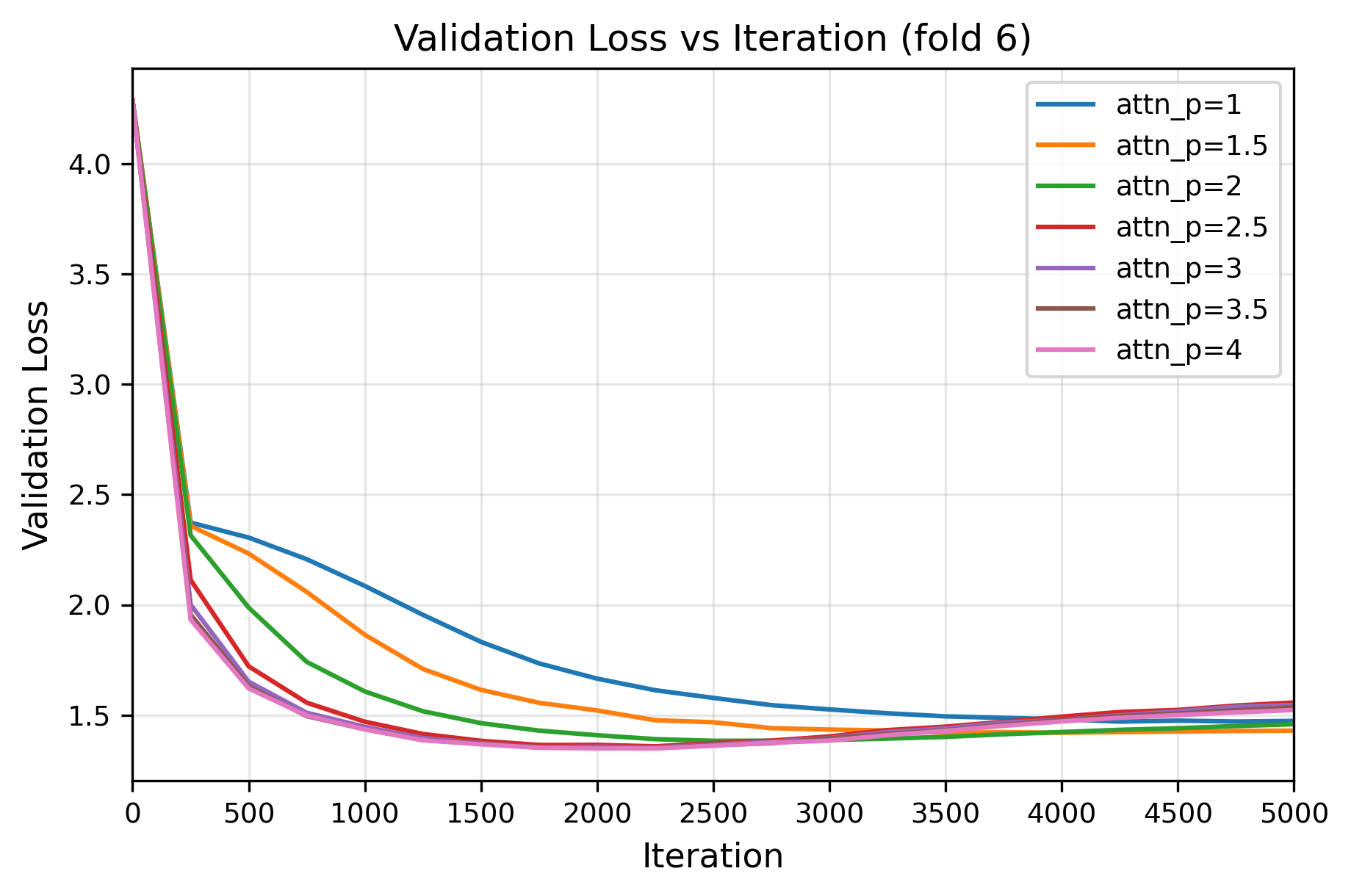}}{\fbox{\scriptsize Missing: fold\_6}}
    {\scriptsize Fold 6}
  \end{minipage}\hfill
  \begin{minipage}[t]{0.24\linewidth}
    \centering
    \IfFileExists{images/fold_7_attn_p_comparison.png}{\includegraphics[width=\linewidth]{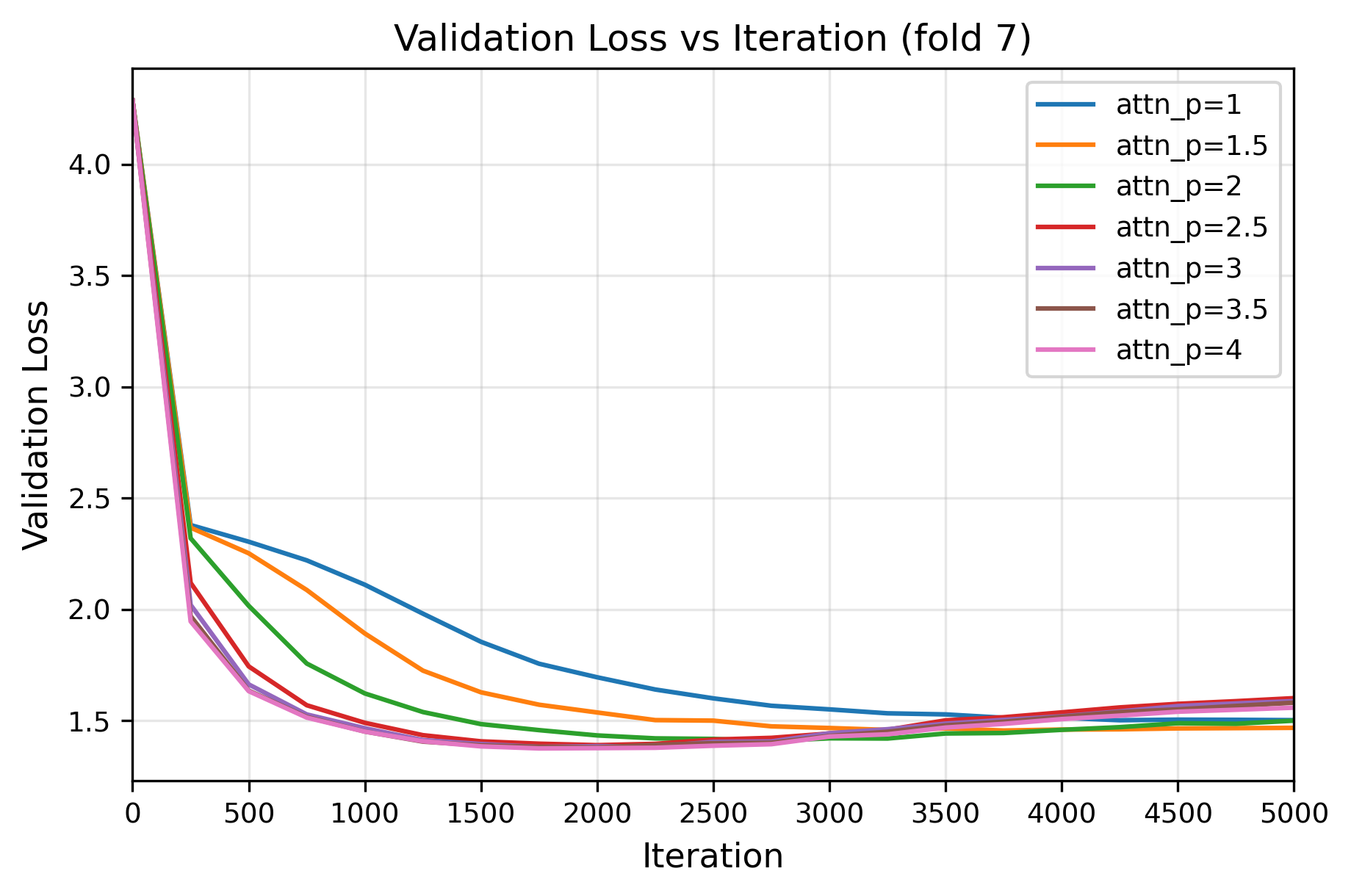}}{\fbox{\scriptsize Missing: fold\_7}}
    {\scriptsize Fold 7}
  \end{minipage}

  \vspace{0.8em}

  \hspace*{\fill}
  \begin{minipage}[t]{0.24\linewidth}
    \centering
    \IfFileExists{images/fold_8_attn_p_comparison.png}{\includegraphics[width=\linewidth]{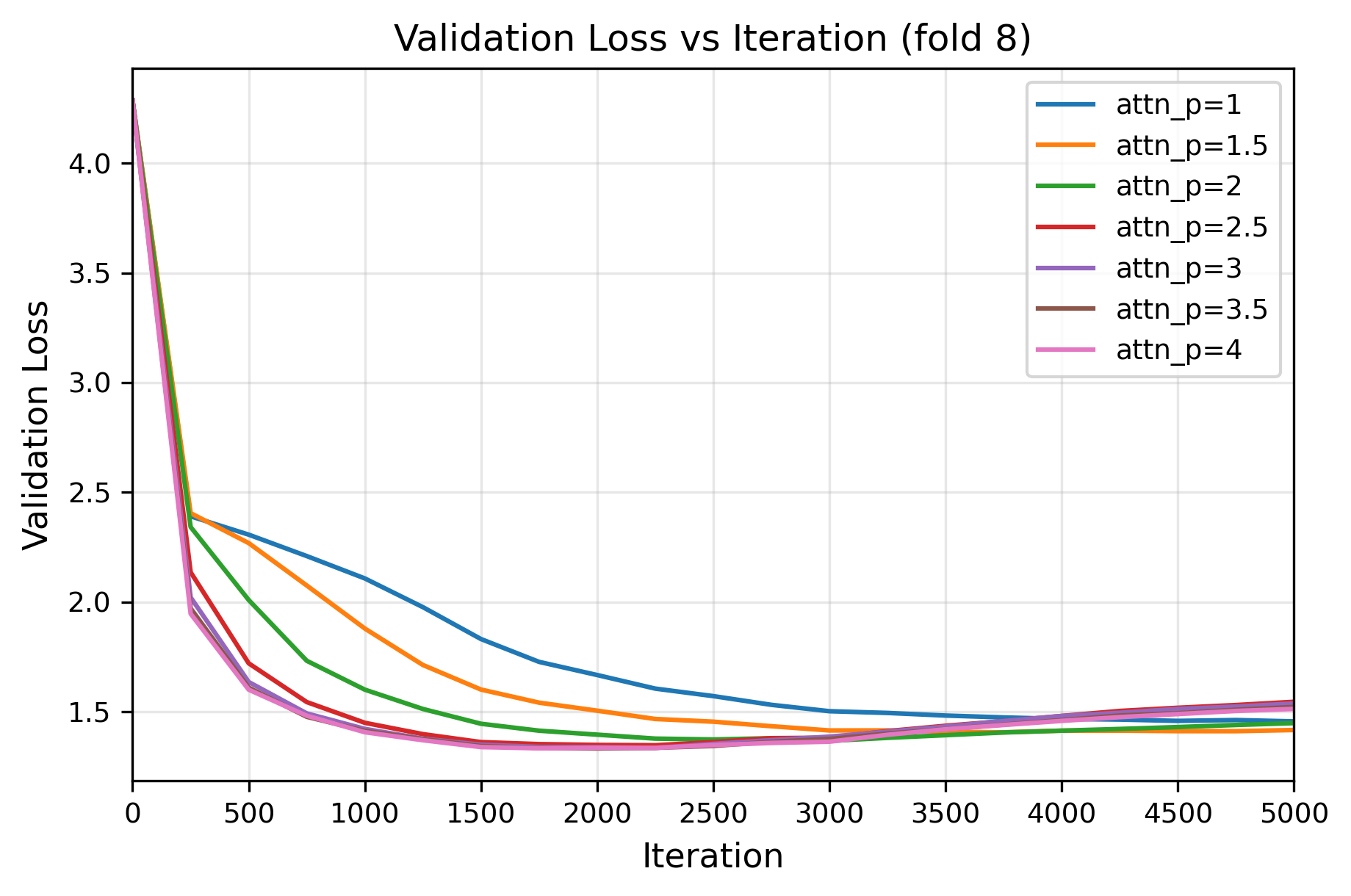}}{\fbox{\scriptsize Missing: fold\_8}}
    {\scriptsize Fold 8}
  \end{minipage}\hfill
  \begin{minipage}[t]{0.24\linewidth}
    \centering
    \IfFileExists{images/fold_9_attn_p_comparison.png}{\includegraphics[width=\linewidth]{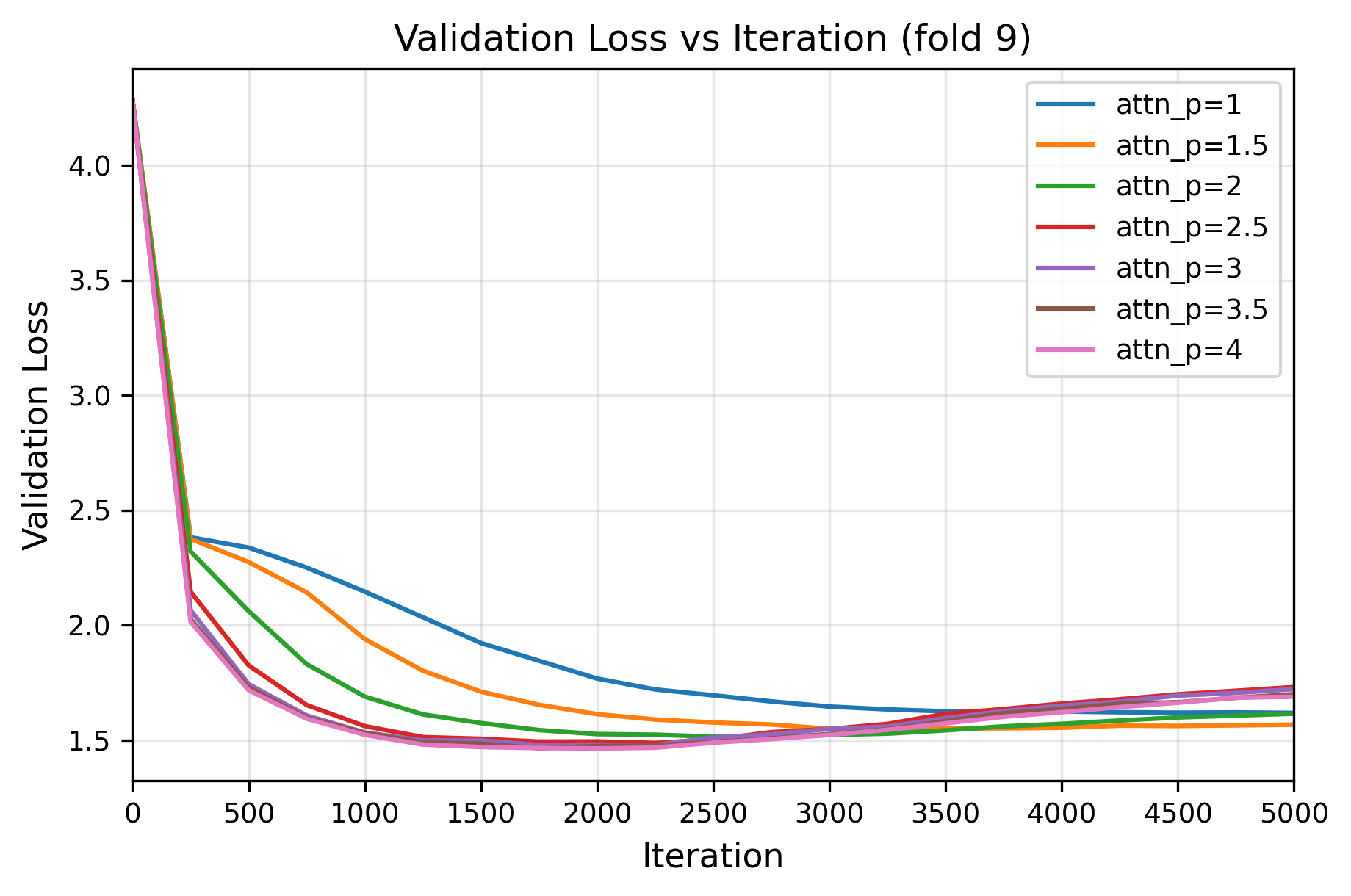}}{\fbox{\scriptsize Missing: fold\_9}}
    {\scriptsize Fold 9}
  \end{minipage}
  \hspace*{\fill}

  \caption{Validation loss vs. iterations for each fold (KFold with $K=10$).}
  \label{fig:perfold_all}
\end{figure}

\section*{Author contributions}
Conceptualization, E.L.-R.; methodology, E.L.-R.; software, M.D.-G.; validation, E.L.-R. and E.J.P.; formal analysis, E.L.-R.; investigation, E.L.-R.; resources, M.D.-G.; data curation, M.D.-G.; writing---original draft preparation, E.L.-R. and M.D.-G.; writing---review and editing, E.L.-R., M.D.-G. and E.J.P.; visualization, M.D.-G.; supervision, E.L.-R.; project administration, E.L.-R.; funding acquisition, E.L.-R. and E.J.P. All authors have read and agreed to the published version of the manuscript.

\bibliographystyle{unsrtnat}
\bibliography{references}  






\end{document}